\documentclass[runningheads]{llncs}
\usepackage{graphicx}
\usepackage{tikz}
\usepackage{comment}
\usepackage{amsmath,amssymb} % define this before the line numbering.
\usepackage{color}

\usepackage{epsfig}
\usepackage{graphicx}
\usepackage{amsmath}
\usepackage{amssymb}
\usepackage{graphicx}
\usepackage{makecell}
\usepackage{epstopdf}
\usepackage{hyperref}					% Reference
\hypersetup{colorlinks,%				% Reference color setup	
	linkcolor=red,%
	citecolor=blue}
\usepackage{amsfonts}
\usepackage{multirow}
\usepackage{setspace}
\usepackage{booktabs}
\usepackage{enumerate}
\usepackage[linesnumbered,ruled,vlined]{algorithm2e}
\usepackage{mathrsfs}
\usepackage{setspace}
\usepackage{colortbl} 
\usepackage{arydshln}

\usepackage[accsupp]{axessibility}  % Improves PDF readability for those with disabilities.
\definecolor{mygray}{gray}{.95}
\usepackage{orcidlink}
\begin{document}
	% \linenumbers
	\pagestyle{headings}
	\mainmatter
	\def\ECCVSubNumber{4640}  % Insert your submission number here
	
	\title{Self-supervised Social Relation Representation for Human Group Detection} % Replace with your title
	
	\begin{comment}
	\titlerunning{ECCV-22 submission ID \ECCVSubNumber} 
	\authorrunning{ECCV-22 submission ID \ECCVSubNumber} 
	\author{Anonymous ECCV submission}
	\institute{Paper ID \ECCVSubNumber}
	\end{comment}
	%******************

	%\begin{comment}
	\titlerunning{Self-supervised Social Relation Representation for Human Group Detection}
	\author{Jiacheng Li\inst{1}$^*$\orcidlink{0000-0002-2078-5998} \and
		Ruize Han\inst{1}$^*$$^\dagger$\orcidlink{0000-0002-6587-8936} \and
		Haomin Yan\inst{1} \and 
		Zekun Qian \inst{1} 
		\and \\ Wei Feng \inst{1}\orcidlink{0000-0003-3809-1086}
		\and Song Wang \inst{2}\orcidlink{0000-0003-4152-5295}}
	\authorrunning{J. Li, R. Han et al.}
	\institute{Intelligence and Computing College, Tianjin University, China \and
		University of South Carolina, Columbia, USA \\
	\email{\{threeswords, han\_ruize, yan\_hm, clarkqian, wfeng\}@tju.edu.cn, songwang@cec.sc.edu }}
%		\email{${\dagger}$ han\_ruize@tju.edu.cn } }
	%\end{comment}
	%******************

	\maketitle
	
	\newcommand\blfootnote[1]{%
		\begingroup
		\renewcommand\thefootnote{}\footnote{#1}%
		\addtocounter{footnote}{-1}%
		\endgroup
	}
	
	\begin{abstract}
		
		%Human group detection is a fundamental task for crowded scene analysis, which has many real-world applications, e.g., video surveillance.
		Human group detection, which splits crowd of people into groups, is an important step for video-based human social activity analysis.
		The core of human group detection is the human social relation representation and division.
		In this paper, we propose a new two-stage multi-head framework for human group detection. 
		In the first stage, we propose a human behavior simulator head to learn the social relation feature embedding, which is self-supervised trained by leveraging the socially grounded multi-person behavior relationship.
		In the second stage, based on the social relation embedding, we develop a self-attention inspired network for human group detection.
		Remarkable performance on two state-of-the-art large-scale benchmarks, i.e., PANDA and JRDB-Group, verifies the effectiveness of the proposed framework. Benefiting from the self-supervised social relation embedding, our method can provide promising results with very few (labeled) training data. We have released the source code to the public.
		
	\end{abstract}
	
	\section{Introduction}
	
	% Background & Importance
	
	%% group in the Crowd
	%% 
	%	People crowd is a very common scene in real world, e.g., in outdoor parade, social gatherings. The analysis of people crowd scene is a significant task in computer vision, which has many application such as video surveillance, social relation understanding.
	%	Recently, many researches on people crowd analysis focus on the human counting, pedestrian detection and tracking, and group activity recognition, etc. These tasks handle the people crowd either in the individual level or in the global level.
	%	Actually, human groups are recognized as the basic elements composing the crowd~\cite{moussaid2010walking}. Thus, the human group detection, i.e., identifying groups of people
	%	from crowds, is an important task, which, however, is not well studied in recent works.
	
	\blfootnote{$^*$Equal Contribution. $^\dagger$Corresponding Author.}Scenes with a crowd are very common in the real world, e.g., outdoor parades and social gatherings~\cite{Han_2022_CVPR,han2022multiview}. Such scenes lead to many important video-surveillance tasks, such as human counting, pedestrian tracking~\cite{han2020complementary,han2021multiple,gan2021self}, and group activity recognition~\cite{zhao2020human}. These tasks involve crowd analysis either locally for individuals or more globally for the whole crowd. With the use of wide-view cameras in surveillance, we can now collect videos to cover a large-scale crowd composed of multiple groups of people with high resolution. In this case, we usually need to first identify all these groups~\cite{moussaid2010walking} before conducting further human activity analysis. This leads to the important task of (video-based) human group detection, which tries to divide the crowd into multiple non-overlapped human groups.
	
	\begin{figure}[ht!] 
		\centering 
		\includegraphics[width=1\linewidth]{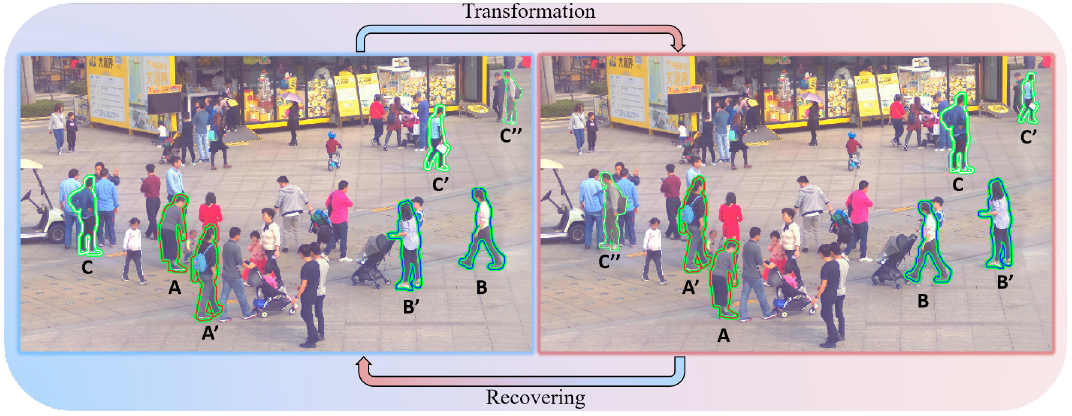}
		\caption{A conceptual illustration of the rationale of our insight. We swap some people in the scene and try to discover the unreasonable states in the transformed scene (right) by implicitly learning the human social relations. Specifically, in the left image, we swap person $A$ with $A'$, $B$ with $B'$ and $C,C',C''$ with $C',C'',C$, respectively, resulting in the right image. Such swapped results in the right contain persons with abnormal behaviors. To recover the above augmentation, the method needs to understand the reasonable state of the person in the crowd considering itself and its surrounding people, which can be used for the social relation representation.}
		\label{fig:insp} % \vspace{-15pt}
	\end{figure}
	
	The human group detection task is very challenging since groups can be formed in complicated ways~\cite{lerner2007crowds,solera2015socially}, whose characteristic is more complex than the pedestrians acting alone or the crowds as a whole. Previous works for group detection mainly leverage the features extracted from either the pedestrian trajectories~\cite{pellegrini2009you} or local spatial structure~\cite{choi2014discovering}, followed by a similarity measurement and clustering analysis, e.g., the weighted graph model~\cite{chang2011probabilistic}, potentially infinite mixture model~\cite{ge2012vision} and correlation clustering~\cite{solera2015socially}.
	In a recent work of PANDA~\cite{wang2020panda}, a new video benchmark that uses gigapixel-level cameras for capturing a supersized scene with hundreds of people was developed, together with an official baseline method that combines the global trajectory and local interaction information for group detection.
	% Motivation & Challenge  & related work
	%%  
	%% hard to annotate
	%% self-supervised method
	
	However, the above methods commonly have two weaknesses. First, the features used in most of the previous group detection methods are pre-defined, many are actually hand-crafted features, including temporal trajectories and spatial distances~\cite{pellegrini2010improving,yamaguchi2011you,choi2014discovering,solera2015socially}. 
	Second, previous works based on machine learning are usually fully supervised and require large-scale annotations~\cite{yamaguchi2011you,wang2020panda}. Nevertheless, the human group detection labels in a dense crowd are very laborious and costly to annotate manually.
	
	% Solution & Contribution
	To address these problems, in this paper, we study spontaneous learning of human relation embedding features for human group detection.
	This is inspired by the sociological interpretation of group -- two or more people interacting to reach a common goal and perceiving a shared membership, based on both physical identity (spatial proximity) and social identity (intra-group rules)~\cite{turner2010towards}.
	For a person in the crowd scene, her/his activity and motion characteristics are highly influenced by the nearby people, especially the interactive ones. 
	For example, as shown in Figure~\ref{fig:insp}, we swap the spatial positions of some people (left and right images are the ones before and after the swapping). We can easily find that the `transformed' people in the right image show unreasonable behaviors, e.g., in the right of Figure~\ref{fig:insp}, the strange pose (A, B'), dangerous moving direction with close distance to something (A', B).
	Such visual unreasonability, in the human sense, is dependent on the human-human interactive behavior and social relation, which are very important for human group detection. 
	Inspired by this rationale, we try to develop a method to automatically learn human social relations through such unreasonable behaviors.	
	More specifically, we design a rational-behavior simulator head (BS-Head) to discover the swapped persons and recover their original states based on the information of the surrounding persons.
	This head is used as a self-supervised training head to learn the human social relation embedding. 
	After that, we use the embedding feature produced by BS-Head to train another self-attention-based group prediction network for human group detection. 	
	The main contributions in this work are:
	\begin{itemize}
		\item We propose a two-branch group detection framework, which contains a shared relation embedding module and two heads for two-stage training. It is effective for social group detection problems in large-scale crowds.
		\item We develop a self-supervised relation embedding training strategy, which models the socially grounded human social relations in the crowd scene as the pretext task. The self-supervised representation can benefit the network training using only very few (labeled) data. 	
		\item Experiment results on two newly proposed large-scale benchmarks, i.e., PANDA and JRDB-Group, verify the effectiveness of the proposed method. 
		The proposed method achieves significant group detection performance improvement compared to the state-of-the-art baselines -- \textit{20.7 $\rightarrow$ 53.2\%} and \textit{31.4\% $\rightarrow$ 56.9\%} in F$_1$ score on PANDA and JRDB-Group, respectively.
		We release the source code to the public at~\href {https://github.com/Jiaoma/SHGD}{\textcolor{magenta}{https://github.com/Jiaoma/SHGD}}.
	\end{itemize}

	\section{Related Work}
	
	\textbf{Human group detection.}
	Human group detection is an essential task in computer vision, which, however, has not been widely studied recently.
	Early human group detection methods can be divided into three categories, including group-based methods, individual-group methods, and individual-based methods.
	In group-based methods, no individual information is considered~\cite{shao2014scene,feldmann2010tracking}. The individual-group methods try to integrate the information of human trajectories instead of separately using each one~\cite{pang2011detection,bazzani2012decentralized}. The individual-based methods consider individual subjects, leading to many models, such as predicting the social group relations by using the weighted graph~\cite{chang2011probabilistic}, potentially infinite mixture model~\cite{ge2012vision} and correlation clustering~\cite{solera2015socially}, etc.
	Recently, Shao et al.~\cite{shao2018real} employed the goal directions instead of traditional positions and velocities to find group members. 
	%In~\cite{fernando2018gd} attempt to remove the supervised learning requirement for group detection and use augmented context embedding to learn the group attributes automatically.
	More recently, instead of focusing only on the position-aware group detection, PANDA~\cite{wang2020panda} proposes to integrate the dynamic human interactions for human group detection.	
	As mentioned earlier, these methods either use pre-defined, usually hand-crafted, features~\cite{pellegrini2010improving,yamaguchi2011you,choi2014discovering,solera2015socially} or require large-scale annotated data~\cite{yamaguchi2011you,wang2020panda} for training.
	%	There are two problems in the above approaches: 1) The features are pre-defined, especially the hand-draft features such as the temporal trajectories or spatial distances~\cite{pellegrini2010improving,yamaguchi2011you,choi2014discovering,solera2015socially}. 2) The learning based methods require the group identification labels for feature extraction~\cite{yamaguchi2011you,wang2020panda}, which, however, are very difficult to annotate manually.
	
	%\textbf{Group detection datasets.}
	
	\textbf{Human relation discovery.} A key problem for the group detection task is the discovery of human relations, especially social relationship~\cite{li2020graph,Goel_2019_CVPR}, e.g., friends and colleagues, via various human attributes, e.g., age, job, etc.
	Other relations include the human-object interaction (HOI)~\cite{Zhou_2020_CVPR,Zou_2021_CVPR}, human-human interaction (HHI)~\cite{monfort2019moments,zhao2020human}, and gaze communication~\cite{Fan_2019_ICCV}, etc.	
	For HOI or HHI, relations are usually specific and mostly determined by the involved subjects/objects, e.g., a scene with a human and a bike is more likely to be detected as `ride bike'. 
	%	That is to say, the relation category between the subjects has commonly been priorly defined, e.g., `reading book` for HOI, `shaking hand` for HHI, etc.
	Differently, our task explores the relations without the above specific prior knowledge, e.g., the social relationships among people in the crowd. 
	
	\textbf{Human relation modeling in the multi-person scene.} 
	Group-wise human relation is also collaboratively or implicitly used in many other related tasks, and human group detection is often accompanied by pedestrian trajectory prediction~\cite{pellegrini2010improving,yamaguchi2011you,fernando2018gd}. The method in~\cite{pellegrini2010improving} tries to jointly achieve the pedestrian trajectory prediction and group membership estimation using a Conditional Random Field (CRF) model. Similarly, an SVM-based method is proposed in~\cite{yamaguchi2011you} to handle the estimations of pedestrian trajectory and social (group) relationships. A GAN pipeline is proposed in~\cite{shao2018real} that learns informative latent features for joint pedestrian trajectory forecasting and group detection. Social-GAN~\cite{gupta2018social} proposes to predict the socially-accepted motion trajectories in crowded scenes by considering human social relations. Multi-human relations have also been considered in group activity recognition (GAR)~\cite{GAR-ARG,dataset-NBA,SSU,AT,pramono2020empowering,yuan2021learning}. 
	For example, ARG~\cite{GAR-ARG} proposes to build a flexible and efficient actor relation graph to simultaneously capture the appearance and position relation between actors. The method in~\cite{pramono2020empowering} considers multiple cliques with different scales of the locality to account for the diversity of the actors’ relations in group activities.
	The above methods all take the human relation modeling or detection as an auxiliary task but do not divide the crowd into groups that is our focus in this paper.
	
	%	\textbf{Casual discovery}
	%	Adapting Neural Networks for the Estimation of
	%	Treatment Effects
	%	
	%	Causal Inference via Sparse Additive Models
	%	with Application to Online Advertising
	%	
	%	Improving Causal Inference by Increasing Model Expressiveness

	\section{Proposed Method}

	\subsection{Overview}
	We first give an overview of the proposed method.
	{Given a video clip recording a crowded multi-person scene, we denote $N$ as the maximum number of people (referred to as subject in this paper) within $T$ frames\footnote[1]{If a subject is missing in a frame, we fill it with blank (all-zero feature vector). 
%	Here we use the object tracklet provided in the dataset.
}, e.g., 10 frames, where $N$ could be very large, e.g., over 1,000 in PANDA dataset.}
	{We first extract a feature encoding vector $\textbf{v}^t_{i} \in \mathbb{R}^{C_v}$ for the $i$-{th} subject in $t$-{th} frame, where $C_v$ is the number of feature dimensions. The encoding vector can be represented by both the human skeleton joints and spatial location.}
	{We model all the the subjects in the scene as an undirected graph $\mathcal{G}$ = $<\textbf{V}, \textbf{R}>$, in which $\textbf{V} \in \mathbb{R}^{T\times N \times {C_v}} $ denotes the set for all $\textbf{v}^t_{i}$, and $\textbf{R} \in \mathbb{R}^{T \times  N^2}$ encodes the \textit{group relation} $r^t_{i,j}$ between subject $i$ and subject $j$ at frame $t$. It is in the range of $[0,1]$, where $1$ means the subject $j$ and subject $i$ belong to the same group.}
	{The relation matrix $\textbf{R}$ can be used to divide the crowd of people into groups by label propagation methods~\cite{zhan2018consensus}.}
	
%	\vspace{-10pt}
	\begin{figure*}[ht!] 
		\centering 
		\includegraphics[width=0.95\linewidth]{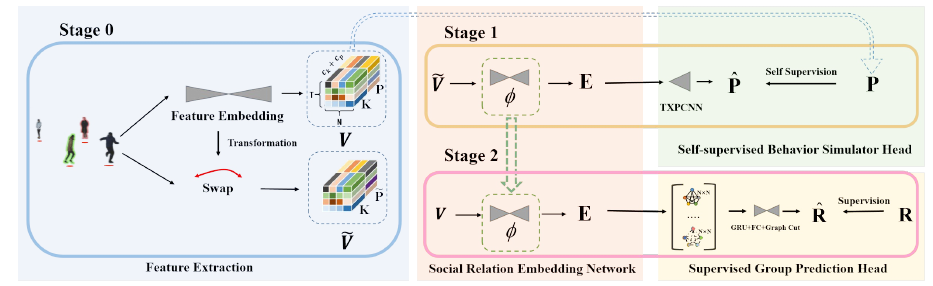} % \vspace{-20pt}
		\caption{An illustration of the overall framework of our method, which is composed of an offline feature extraction module and a social relation embedding module for feature construction, followed by two heads, i.e., the behavior simulator head (BS-Head) and the group prediction head. The overall network is trained in two stages. In stage 1, we train the relation embedding network with the BS-Head in a self-supervised manner. In stage 2, based on the relation embedding features obtained by stage 1, we train our network with the group prediction head for generating the group detection results. }
		\label{fig:framework} % \vspace{-10pt}
	\end{figure*}
	%We also define an undirected graph $\mathcal{U}^{t}=<\mathbf{V}^{t}, \mathbf{S}^{t}>$, which is a graph connecting each person with their euclidean distance, i.e., $s^{t}_{i,j} \in \textbf{S}^{t}$,
	%\begin{equation}
	%s^{t}_{i,j}=s^{t}_{j,i}={||(x_i,y_i)-(x_j,y_j)||}_2
	%\end{equation} 
	%where  $(x_i,y_i)$ denotes the spatial coordinate of person $i$.
	%With above definition, we assume that $\textbf{v}^{t}_{i}$ is possibly influenced by some of its neighbored persons $P^{t}_{q}$ in the set $\{P^{t}_{q}| q \in \mathcal{N}_i \}$, in which $\mathcal{N}_i$ is the set of persons that have distance no more than a threshold, i.e., a possible human-influence range. 	Also, $\textbf{v}^{t}_{i}$ could be not influenced by any other ones (i.e., there is no surrounding people around), this way we have  $\mathcal{N}_i = \emptyset$. 
	{To obtain such relation matrix $\textbf{R}$, many deep learning methods~\cite{thompson2021conversational,swofford2020improving} generally first encode the features $\textbf{V}$ into a group relation embedding space and then predict $\textbf{R}$ using a relation modeling algorithm, like the graph neural networks (GNN).}
	{In this paper, we aim at developing an effective group relation embedding method in a self-supervised manner, and then use such embedding $\textbf{E}$ to predict the relation matrix ${\textbf{R}}$.}{
		The pipeline of the proposed method could be summarized as two stages: 1) We first train a real-world surrounding-subject-aware \textit{human behavior simulator} in a self-supervised manner, which is to spontaneously learn the natural human behavior within a group and embed the features into a relation embedding space. 2) Based on the relation embedding, we then learn a group prediction sub-network, which is trained for discovering the human-human group relations among multiple persons. This stage can only use a small number of labeled data to train.%3) We finally fix the parameters of the relation network and add a simple fully-connected (FC) layer based network to achieve the human group detection task.
	}
	{The architecture of the whole network is shown in Figure~\ref{fig:framework}. It consists of two prediction heads that share the same feature extraction and social relation embedding network. The whole network uses the human detection bounding boxes of the $N$ persons each frame in $T$ frames as input for feature construction, as discussed in Section~\ref{sec:net}. It has two kinds of outputs produced by its two heads separately in the two-stage training, as presented in Section~\ref{sec:simulator} and Section~\ref{sec:group}, respectively.

		\subsection{Feature Construction Module}
		\label{sec:net}
		{\textbf{Social relation embedding network.} 	
%		\textbf{Feature extraction.} 
		The original feature $\textbf{V} \in \mathbb{R}^{T \times N\times C_v}$ is constructed by concatenating both the 2D skeleton joints $\textbf{K} \in \mathbb{R}^{T\times N \times C_k}$ and the positional feature $\textbf{P} \in \mathbb{R}^{T\times N \times C_p} $ of all $N$ subjects at frame $t$.%, where the 2D skeleton joints are detected by the Unipose network~\cite{unipose} and 2D positional (human bounding box center) feature is constructed into higher dimension by a positional encoding method as in~\cite{vaswani2017attention}.}

	The social relation embedding network uses $\textbf{V}$ as input and outputs a social relation embedding feature $\textbf{E}$, i.e., 
	\begin{equation}
		\textbf{E} = \phi(\textbf{V}) \in \mathbb{R}^{T\times N\times C_e}.
	\end{equation}
	Specifically, first, for the skeleton feature $\textbf{K}$, we adopt the Shift-GCN~\cite{shiftgcn} alike network to extract the temporal action features from the human skeleton joint. Also, for the positional feature $\textbf{P}$, we concatenate it with the skeleton feature $\textbf{K}$ and utilize the architecture in spatial-temporal graph convolutional network (ST-GCNN)~\cite{mohamed2020social} to model the spatial and temporal information. 
%	which produces the pose-position feature block $\textbf{E}$ as the social relation embedding of whole scene.}}

	{\subsection{Self-supervised Behavior Simulator Head}\label{sec:simulator}
	
	In this section, we present the rationale and process to train the social relation embedding network in a self-supervised manner with the proposed rational-behavior simulator head (BS-Head).
	Specifically, we aim to establish a human behavior simulator to model the \textit{reasonable behavior} of the humans in a crowd with the constraint of its social relation/rules/etiquette with other people, e.g., the greeting from surrounding people and the relative position distribution to the others.
	Our basic assumption is that every person should behave logically and not do unreasonable things like talking to nobody, stepping over others, totally ignoring others' greetings, etc. This way, we assume that a person with its surrounding subjects can be modeled by an implicit distribution $D$.
	Specifically, given a person $c$ and its neighbors (with closer distance), we have
	\begin{equation}
		\label{eq:H}
		H_c \triangleq (\textbf{V}_c,\textbf{V}_{n}) \sim D, n \in \mathcal{N}_c,
	\end{equation}
	where $\textbf{V}_c$ is the extracted feature of the center subject $c$, and $\textbf{V}_{n}$ with $n \in \mathcal{N}_c$ denotes the features of the neighboring subjects, which are identified by the spatial distance (in the image) to $c$ within a threshold.
	We define $H_c$ as the \textit{neighbor feature cluster} by pairing the center subject and surrounding subjects. 
%	 \vspace{-15pt}	
	
	To learn such distribution, we propose a self-supervised learning method.
	Our basic idea is that if the state (e.g., spatial position, human behavior) of a center subject is artificially changed, the states of its surrounding subjects can synergistically guide the center subject to be recovered to its original state.
	%	Specifically, we first sample a real case $H_i \sim D$. 
	This way, as shown in Figure~\ref{fig:feat}, we first destroy the state of the feature block ${\textbf{V}}$ and vary it by {spatial-temporal transformation}, i.e., swapping its position with another one, as shown in Figure~\ref{fig:insp}. We can perform the swapping directly in the feature space, to get $\tilde{\textbf{V}}$ as
	\begin{equation}
		\tilde{\textbf{V}} = \chi(\textbf{V}) \in \mathbb{R}^{N\times C_v},
	\end{equation}
	where $\chi$ denotes the transformation operation. 
	
	\begin{figure*}[ht!] 
		\centering 
		\includegraphics[width=0.65\linewidth]{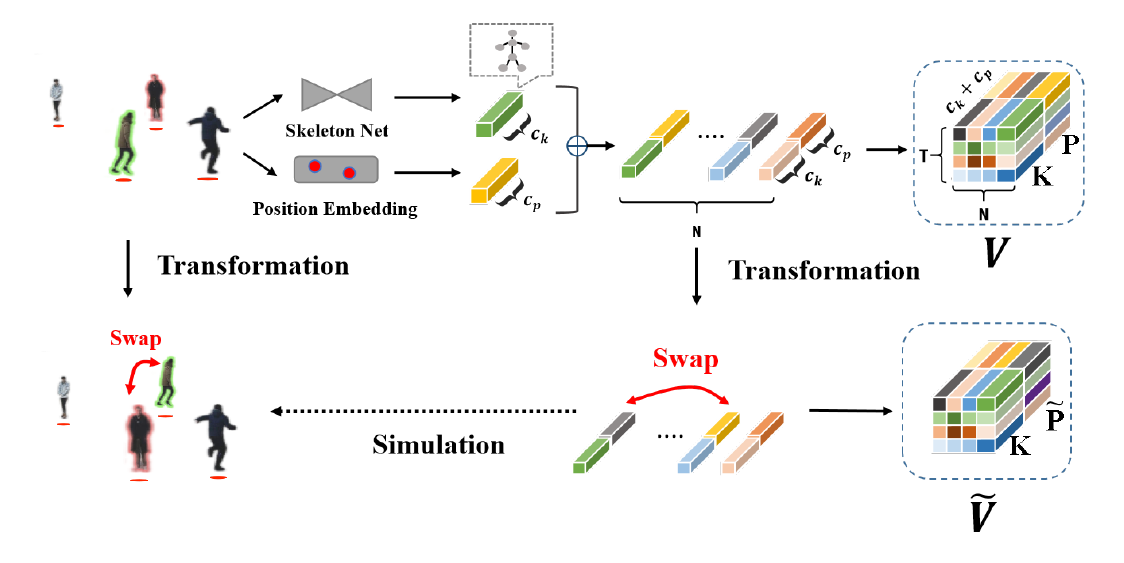} % \vspace{-10pt}
		\caption{An illustration of the feature transformation. }
		\label{fig:feat} % \vspace{-15pt}
	\end{figure*}
	
	The transformed feature $\tilde{\textbf{V}}$ and its surrounding subjects are then fed to the BS-Head, which tries to recover the original feature  $\textbf{V}$. 
	This way, with the BS-Head, the training process of the first-stage network can be implemented as
	\begin{equation}
		\label{eq:BS}
		\hat{\textbf{V}}  = \mathrm{BSHead}(\phi(\tilde{\textbf{V}}))  \to \textbf{V},
	\end{equation}
	where $\mathrm{BSHead}$ denotes the rational-behavior simulator head. It tries to generate a recovered feature $\hat{\textbf{V}}$ tending to approximate the original feature $\textbf{V}$.
	Here $\phi$ denotes the social relation embedding network as discussed in Section~\ref{sec:net}.
	For the behavior simulator head (BS-Head), we adopt the structure in the graph convolution neural network TXP-CNN~\cite{mohamed2020social}. 	
	
		\textbf{Discussion.}
		As shown in the left of Figure~\ref{fig:BS}, the transformed feature $\tilde{\textbf{V}} \in \mathbb{R}^{T\times N\times C_v}$ is constructed by the skeleton feature $\textbf{K}$ and the positional feature $\tilde{\textbf{P}}$, where the transformation operation $\chi$ is only applied to the positional feature $\textbf{P}$, i.e., we simulate to exchange the position of a subject with another while maintaining its skeleton (representing the action and pose of a human). This may make the subject with original behavior in the swapped position look unreasonable with its (new) neighbors (see subjects A and B' in the right of Figure~\ref{fig:insp}). Also, the $\mathrm{BSHead}$ actually only outputs the recovered positional feature $\hat{\textbf{P}}$ (not including the skeleton feature), which is self-supervised by that from the original feature block ${\textbf{P}}$, as shown in Figure~\ref{fig:framework}. 
		%	It takes a sequence of $T\times N$ (frames $\times$ subjects/frame) social relation embedding $\textbf{E} $ together as input and correspondingly generates $T\times N$ outputs $\hat{\textbf{P}}$.}		
		%	 \vspace{-10pt}
		\begin{figure*}[ht!] 
			\centering 
			\includegraphics[width=0.825\linewidth]{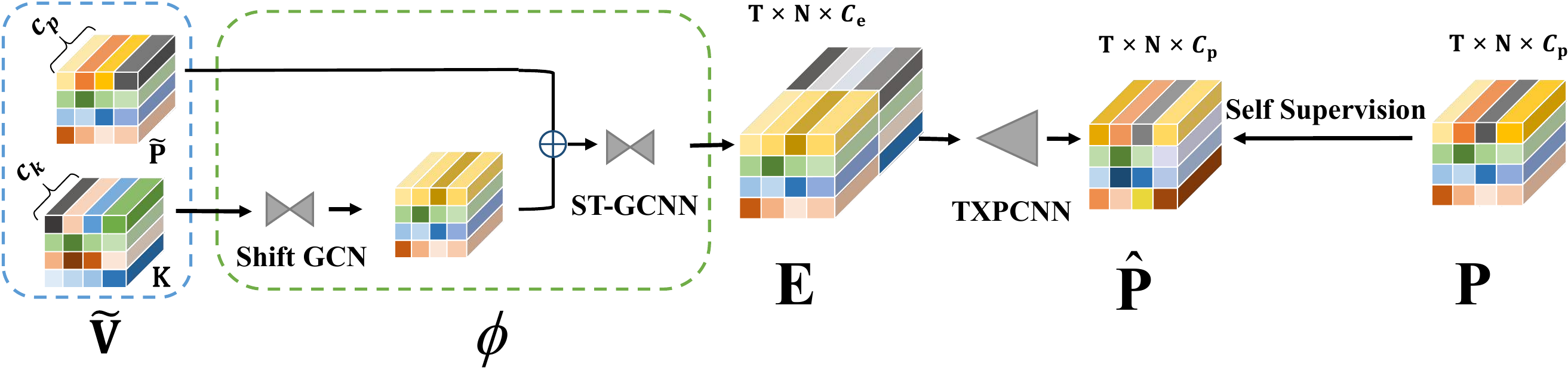}%  \vspace{-10pt}
			\caption{An illustration of the self-supervised behavior simulator head, which trains the relation embedding network with the BS-Head in a self-supervised manner.}
			\label{fig:BS} % \vspace{-10pt}
		\end{figure*}
		%
		% and a adjacent matrix $\mathcal{A} \in \mathcal{R}^{T\times N\times N}$. Here $T$ is the number of frames, $N$ is the maximum number of persons in one frame during the $T$ frames, $C$ is the number of feature channels, plus $1$ in channel due to the mask we use as described in Section~\ref{sec:simulator}. In this work, we employ this network to output a recovered node embedding matrix $\tilde{\mathcal{M}_1} \in \mathcal{R}^{T\times N\times (C+1)}$ of the same time period as $\mathcal{M}_1$. 		
		\subsection{Supervised Group Prediction Head}
		\label{sec:group}
		The above behavior simulator head is mainly used for training the social relation embedding network $\phi$ in a self-supervised manner, which, however, can not provide the group detection results.
		We propose the group detection head, as shown in Figure~\ref{fig:GPD}, which consists of a stacked attention module (Stack-Att), a GRU, and a linear layer to predict the human relation matrix, which will be post-processed by a label propagation method~\cite{zhan2018consensus} to generate the predicted human group results. 
		
		\begin{figure*}[h!] 
			\centering 
			\includegraphics[width=0.825\linewidth]{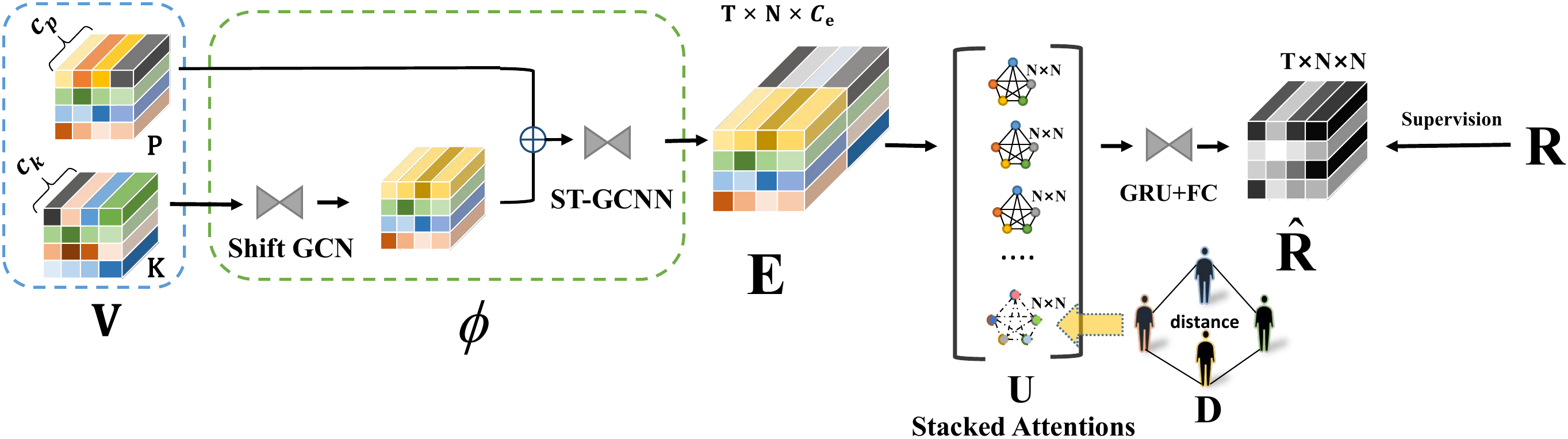} %\vspace{-10pt}
			\caption{An illustration of the second stage of our method, based on the relation embedding features obtained by the first stage, which trains the network with the group prediction head for generating the final group detection results. }
			\label{fig:GPD} %\vspace{-10pt}
		\end{figure*}
		
		This head takes the original feature block $\textbf{V}$ as input, which is first fed into the embedding network $\phi$ (trained by the first stage) to get social relation embedding $\textbf{E} \in \mathbb{R}^{T\times N\times Ce}$. 
		We then apply a stacked attention modules (Stack-Att), which uses $M$ (a pre-set parameter) independent attention module operating separately as
		\begin{equation}
			\mathrm{att}^m(\mathbf{e}_i, e_j)=\frac{\theta({\mathbf{e}_i})^T\theta({\mathbf{e}_j})}{\sqrt{L}} \in \mathbb{R}, \quad m=1,2,\cdots,M,
		\end{equation}
		where  $\mathbf{e}_i  \in \mathbb{R}^{C_e}$ denotes an element of $\textbf{E}$, and $\theta({\mathbf{e}_i})=\mathbf{W}_{\theta}{\mathbf{e}_i}+\mathbf{b}_{\theta}$ is the learnable linear transformations, $L$ denotes the vector length of $\theta({\mathbf{e}_i})$. 
		For each frame containing $N$ subjects, we generate $N \times N$ pairs of elements $(i,j)$.
		% $\mathbf{W}_{\theta} \in \mathcal{R}^{C_4\times C_5}$ and $\mathbf{b}_{\theta} \in \mathcal{R}^{C_5}$ 
		For all $T\times N \times N$ element pairs $(i,j)$ in $T$ frames, we get the pair-wise relation map $ \textbf{A}^m \in \mathbb{R}^{T\times N\times N}$ with the value ${A}^m_{(i,j)} = \mathrm{att}^m(\mathbf{e}_i, \mathbf{e}_j)$, calculated by the attention module.
		Then we stack all the relation map from $M$ attention modules for all $T$ frames to be a \textit{stacked attention block} $\textbf{U} \in \mathbb{R}^{T\times N\times N \times M}$ as a multi-channel relation feature block. 
		Besides, considering the camera view in this problem is often very wide, and the persons with far distance couldn't be in one group, we also stack a distance matrix $\mathbf{D} \in \mathbb{R}^{T\times N\times N \times 1}$ on $\textbf{U}$ to get the final relation block $\mathbf{U}_D \in \mathbb{R}^{T \times N \times N \times (M+1)}$.
		This block is then fed to a single layer GRU temporal network to aggregate temporal information and a fully-connected layer (FC) to predict the final group relation matrix $\hat{\textbf{R}} \in \mathbb{R}^{T\times N\times N}$, which is supervised by the annotated relation matrix  ${\textbf{R}}$, as shown in Figure~\ref{fig:GPD}.
		
		\textbf{Discussion.} Inspired by the self-attention mechanism in~\cite{vaswani2017attention}~\cite{GAR-ARG}, the proposed Stack-Att adopts the self-attention for each pair of embedding features, i.e., $\mathbf{e}_i, \mathbf{e}_j$  to calculate their relation representation. Compared to the previous methods for relation (edge) feature representation, e.g., directly concatenating the embedding features, the proposed strategy is more memory-efficient, which can handle the challenge of a large number of people in the crowd scene.
		Besides, the self-attention mechanism can better represent the relation between two embedding features. Our method is also different from \cite{vaswani2017attention}~\cite{GAR-ARG}, which only produce $1$-channel edge attention matrix. We apply it $M$ times independently and stack them as a multi-channel edge feature block.

		\subsection{Implementation Details}
		{\textbf{Input features.} As mentioned in Section~\ref{sec:net}, we use 2D skeleton joints $\textbf{K}$ and positional feature $\textbf{P}$ as input. We use Unipose network trained on MPII~\cite{MPII} dataset to detect the 2D skeleton joints of all the people in each frame. The detected 2D skeleton joints $\textbf{k}^{t}_i \in \mathbb{R}^{32}$, for $i=1,2,\cdots,N, t = 1,2,\cdots, T$ has $16$ joints, each of which is a 2D coordinate in the pixel coordinate system. We also use the positional encoding method~\cite{vaswani2017attention} to project the 2D position of each subject's center to the $32$-dimension space as the positional feature $\textbf{P}$. 
			%The distance matrix $\mathbf{D}^{t} \in \mathcal{R}^{N\times N}$ is calculated by ${d}^{t}_{i,j}=	\frac{1}{\sqrt{(x_i-x_j)^2+(y_i-y_j)^2}}$.	
			
		\textbf{Two-stage training.} In the first stage, we use the self-supervised training strategy as discussed in Section~\ref{sec:simulator} to train the whole network using the relation-behavior simulator head, which outputs $\hat{\textbf{P}}$ and we use MSE as loss criterion to enforce $\hat{\textbf{P}}$ be close to $\textbf{P}$. 
		In the first training stage, we apply the swap operation to the training data. For the PANDA dataset, we first randomly select $10\%$ persons in the scene and shuffle their positions as the swap operation. 
		For JRDB dataset, since some frames only contain less than $10$ persons, the selection ratio is $20\%$. 
		Besides, to make the recovering of swap not equal to copying one's origin position, we add small random noises following the uniform distribution on the swapped positions.
		
		In the second stage, we use the trained social relation embedding network $\phi$ (in the first stage) with only the group prediction head to predict the group relation matrix $\hat{\textbf{R}}$, which is enforced to be close to the ground-truth $\textbf{R}$. Here we flatten the group relation matrix and use the cosine similarity loss as a criterion during training. 
		This stage of training is a supervised training process, but it mainly updates the parameters of the group prediction head. 
		This stage also fine-tunes the parameters of the social relation embedding network $\phi$, which could further boost the performance of the whole framework on the group detection task.
%		We conduct the ablation study to verify the training strategy in the following experiment section. 
		With the first stage self-supervised training for $\phi$, the second stage can use few labeled data for training, which is also verified in the experiments.
				
		We use stochastic gradient descent with Adam optimizer to optimize the parameters. {The training of stage 1 takes 200 epochs with the learning rate of $1 \times 10^{-4}$}. {And the training of stage 2 takes 100 epochs with the learning rate of $7 \times 10^{-3}$. The proposed method is implemented based on the PyTorch framework}.
				
		{\textbf{Network inference.}}
		Given the input features $\textbf{V}$, the inference stage only goes through the group prediction head and outputs the group relation matrix $\hat{\textbf{R}} \in \mathbb{R}^{T\times N\times N}$. Then we use the label propagation method~\cite{zhan2018consensus} to split $N$ persons of each frame into non-overlapped groups, where the number of groups is automatically estimated by the algorithm. The groups only containing one person are dropped.
		
%		{\textbf{Swap operation.}}

		\section{Experiments}
				
				\subsection{Datasets and Metrics}
%				{\bf .} 
%				Several previous datasets are built for group detection tasks, which, however, focus on the scenes with only a small-scale human crowd~\cite{cristani2011social,cabrera2018matchnmingle} or composed of low-resolution videos~\cite{pellegrini2009you,lerner2007crowds}.
%				In our experiment, we use two recent public large-scale benchmarks for performance evaluation.
%				\\
				$\bullet$ \textit{PANDA benchmark.} {We first choose the state-of-the-art dataset PANDA~\cite{wang2020panda} for group detection evaluation. It is a multi-human video dataset covering real-world street scenes with a very wide field-of-view (1 $\mathrm{km}^2$ area) and a very dense crowd in the scene, e.g., 4$k$ subjects in one frame. For the group detection task, it provides $9$ long-term videos in different scenes. The training set has $8$ videos while the testing set has $1$ video, as specified in the benchmark. 		
					The average number of frames and bounding boxes per video in the training set is $2, 713$ and $1, 070.4k$, respectively, while the number of frames and bounding boxes in the testing set is $3, 500$ and $335.2k$, respectively.
					The average number of the human group per video is $144.6$ and $75$ in the training and testing video sets, respectively. The PANDA dataset provides human tracking for the group detection task and their baseline methods with evaluation metrics for comparison.}
				\\
				$\bullet$ \textit{JRDB-Group benchmark.} We also include a new benchmark, i.e., JRDB-Group \cite{ehsanpour2021jrdb}, for performance evaluation. JRDB-Group is built based on the JRDB dataset~\cite{martin2021jrdb}, which is captured by a panoramic camera equipped on a robot walking around in the crowded outdoor/indoor multi-person scenes. JRDB-Group provides the human social group annotation, and we use $20$ training and $7$ testing videos in JRDB-Group in our experiment. 
				Following the setting in JRDB-Group, we evaluate the group detection task on the key frames, which are also sampled every 15 frames, generating $1,419$ training and $404$ testing samples.  
				
%				{\bf Evaluation .} 
				Following the metrics in PANDA benchmark~\cite{wang2020panda}, we use the classical Half metrics~\cite{choi2014discovering} including precision, recall, and F$_1$ scores with group member $IoU > 0.5$ for group detection evaluation.

				\subsection{Results}		
				{\bf Comparison methods.} We find that most group detection methods are obsolete without available source code. We try our best to include more related methods with necessary modifications for comparison.\\
				$\bullet$ \textit{Dis.Mat} + \cite{zhan2018consensus}: We first consider a straightforward method in which we calculate the distance among all subjects in the crowd and apply the label propagation algorithm~\cite{zhan2018consensus} used in our inference stage to get the group division.  \\
				%$\bullet$ \textit{GRU:} Given the features used in the proposed method, we use the GRU to temporally aggregate the features and apply the full-connected layers to obtain the group relations among the subjects.  \\
				$\bullet$ \textit{GNN w GRU}: We apply a graph neural network (GNN) with the features in our method as the node feature to model the group relations among the subjects, in which we also apply a GRU model to integrate the temporal information. \\
				$\bullet$ \textit{ARG} is a state-of-the-art method~\cite{GAR-ARG} for human group activity recognition. ARG trains an Inception-v3 network to extract the appearance features and uses GNN to model the mutual relations among the subjects in the crowd. We use the affinity matrix in GNN as the relation matrix for group membership division.
				\\	
				$\bullet$ \textit{Global-to-local}~\cite{wang2020panda} is the baseline method proposed in PANDA~\cite{wang2020panda}, which applies a global-to-local zoom in framework to validate the incremental effectiveness of local visual clues to global trajectories. More specifically, human entities and their relationships are represented as a graph. Then a global-to-local strategy is applied, and the video interaction scores among subjects are estimated by a spatial-temporal ConvNet. The edges in the graph are merged using label propagation, and the cliques remaining in the graph are the group detection results.
				\\
				$\bullet$ \textit{Joint} and \textit{JRDB-Group} are the baselines in JRDB-Group benchmark~\cite{ehsanpour2021jrdb}. \textit{Joint}~\cite{ehsanpour2020joint} proposes to integrate the human group detection task into the group activity recognition problem. Based on it, \textit{JRDB-Group} adds the human spatial features and number of group constraints as losses for the group detection task.
				
%			
		
%				\vspace{-20pt}
				
				We compare the group detection results of the proposed method and other comparison methods. Table~\ref{tab:panda} shows the comparative results on the PANDA benchmark. We can see that all the comparison methods, including the state-of-the-art baseline method Global-to-local reported in the PANDA benchmark and its variants, generate unsatisfactory results. The comparison method with the best performance, i.e., \textit{ARG}, generates an F$_1$ score of $25.4\%$. The proposed method significantly improves the group detection performance and provides a promising result with an F$_1$ score of $53.2\%$. We can see similar results on JRDB-Group as shown in Table~\ref{tab:jrdb}, in which we can see that the comparison methods can generate relatively better results than in PANDA. We also surprisingly find that the simple baseline methods, i.e., the `distance matrix + \cite{zhan2018consensus}' and the `GNN + GRU', provide not bad performance.
				This is because the number of pedestrians and crowd density in JRDB-Group is lower than those in PANDA. 
				The proposed method provides the best performance on JRDB-Group, which also outperforms the comparison methods by a large margin.
				
%					\vspace{-10pt}
				\begin{table}[htbp]
					\caption{Comparative group detection results on PANDA (\%).}	%\vspace{-10pt}
					\label{tab:panda} 
					\begin{spacing}{1.05}
						\centering
						\renewcommand\tabcolsep{10pt}
						\footnotesize
						\begin{tabular}[*c]{l|ccc}
							\Xhline{1pt}
							%\rowcolor{mygray}  
							{Method} &{Precision} & {Recall} & {F1}\\  \hline \hline
							% \rowcolor{mygray} \hline
							%						\rowcolor{mygray}
							Dis.Mat + \cite{zhan2018consensus}  & 42.9	& 12.0 	& 18.8 \\
							%			Dis. w GRU		& 7.7		& 1.3 	& 2.3 \\
							\rowcolor{mygray}
							GNN w GRU		& 41.9		& 17.3 	& 24.5 \\
							
							ARG~\cite{GAR-ARG}	& 34.9		& 20.0 	&25.4 \\
							%			Social STGCN    	& 0.0		& 0.0 	& 0.0 \\
							\hline 
							\rowcolor{mygray}
							Group-to-local~\cite{wang2020panda}			& 23.7 & 12.0 & 16.0 \\
							%						\rowcolor{mygray}
							Group-to-local w Random~\cite{wang2020panda}	& 24.4 & 13.3 & 17.2 \\
							\rowcolor{mygray}
							Group-to-local w Uncertainty~\cite{wang2020panda}  & 29.3 & 16.0 & 20.7 \\
							\hline   \hline
							%			Ours w  only FC         &57.4 &41.3 &48.1 \\
							%			Ours w  one-stage train.         &43.9 &45.2 &44.5 \\
							%						\rowcolor{mygray}
							Ours   & \textbf{55.9} &\textbf{50.7} & \textbf{53.2} \\
							\Xhline{1pt}            
						\end{tabular}
					\end{spacing} 
				\end{table} %\vspace{-30pt}
				
				\begin{table}[htbp]
					\caption{Comparative group detection results on JRDB-Group (\%).}	%\vspace{-10pt}
					\label{tab:jrdb} 
					\begin{spacing}{1.05}
						\centering
						\renewcommand\tabcolsep{10pt}
						\footnotesize
						\begin{tabular}[*c]{l|ccc}
							\Xhline{1pt}
							%\rowcolor{mygray}  
							{Method} &{Precision} & {Recall} & {F1}\\   \hline \hline
							% \rowcolor{mygray} \hline
							%						\rowcolor{mygray}
							Dis.Mat + \cite{zhan2018consensus} 	&57.3 	&23.5  	&33.4  \\
							%			Dis. w GRU		&49.4	&47.3  	&48.3 \\
							\rowcolor{mygray}						
							GNN w GRU		&43.4 	&28.6  	&34.5  \\
							%						\rowcolor{mygray}
							ARG~\cite{GAR-ARG}	&32.5 		&38.4  	&35.2  \\
							%			Social STGCN    	& 0.0		& 0.0 	& 0.0 \\
							\hline 
							\rowcolor{mygray}
							Joint~\cite{ehsanpour2020joint}  &30.0  &28.4  &29.1  \\
							%						\rowcolor{mygray}
							JRDB-Group~\cite{Ehsanpour2021JRDBActAL}  &39.0  &37.9  &38.4  \\
							\hline   \hline
							%			Ours w  only FC         &50.3 &48.1 &49.2 \\
							%			Ours w  one-stage train. &49.8 &47.8 &48.8 \\
							\rowcolor{mygray}
							Ours			&\textbf{57.7} & \textbf{56.2} & \textbf{56.9} \\
							\Xhline{1pt}            
						\end{tabular}
					\end{spacing} 
				\end{table}
								
				\subsection{Ablation Study}
				
				To verify the effectiveness of our method, we conduct the ablation study as below.
				\\
				$\bullet$ \textbf{Ours w/o self-sup. 1$^{st}$-train.} denotes removing the self-supervised representation training in the first stage in our method, and we directly train the whole network (including the representation embedding module and the group 
				forming module) using the training dataset.
				\\
				$\bullet$ \textbf{Ours w/o self-rep. fine-tune.} denotes removing the representation's fine-tuning in our method, and we fix the parameters in the representation embedding module obtained by the self-supervised first-stage training and only train the group prediction head with the training dataset.
				
				As shown in Table~\ref{tab:abla}, we can first see that the proposed network with direct one-stage training can provide acceptable performance at both datasets, which outperforms the baseline methods significantly with F$_1$ scores of 44.5\% on PANDA and 48.8\% on JRDB-Group, respectively.
				Compared with the full two-stage training version of our method, we can also see that the self-supervised representation in the first-stage training is very effective, which further improves the F$_1$ score to 53.2\% on PANDA and 56.9\% on JRDB-Group, respectively.
				Besides, we find that representation networko  fine-tuning with the training dataset is also useful in our method.
				
%				\vspace{-15pt}
				\begin{table*}[htbp]
					\caption{Ablation study results on PANDA and JRDB-Group datasets (\%).}
					\label{tab:abla} %\vspace{-10pt}
					\begin{spacing}{1.05}
						\centering
						\renewcommand\tabcolsep{6pt}
						\footnotesize
						\begin{tabular}[*c]{l|ccc|ccc}
							\Xhline{1pt}
							%\rowcolor{mygray}  
							&\multicolumn{3}{c|}{PANDA} &	\multicolumn{3}{c}{JRDB-Group} \\ \cline{2-7} 
							\multirow{-2}{30pt}{Method} &{Precision} & {Recall} & {F1}  &{Precision} & {Recall} & {F1}\\  \cline{2-4} \hline
							% \rowcolor{mygray} \hline
							w/o  (self-sup.) 1$^{st}$-train        &48.4	&41.3	&44.6  &49.8 &47.8 &48.8  \\
							\rowcolor{mygray}
							w/o  (self-rep.) fine-tune.        &58.6	&45.3	&51.1   &50.3 &48.1 &49.2 \\
							Ours			& 55.9 & 50.7 & 53.2 	& 57.7 & 56.2 & 56.9 \\
							\Xhline{1pt}            
						\end{tabular}
					\end{spacing} % \vspace{-25pt}
				\end{table*}

				\subsection{In-depth Analysis}
				
				Label annotations for the human group detection in a large-scale scene are laborious. Therefore, we propose the self-supervised relation representation, and further investigate the its performance using different amounts of training data. 
				
				As shown in Table~\ref{tab:traindata}, we can first see that the proposed method without the self-supervised first-stage training provides a poor performance with the reduction of training data. Specifically on the PANDA dataset, using very few training data, i.e.,  10\% of all, the group detection F$_1$ score is only 27.6\%, which is much inferior to that using all training data. For the method with the self-supervised first-stage training but without fine-tuning, although performance drops, the results using 10\% training data are acceptable with the F$_1$ score of 37\% and 43.8\% on PANDA and JRDB-Group, respectively. 
				Similarly, the full version of our method provides very promising results with only 10\% training data on two benchmarks. 
				We can see that, on PANDA benchmark with 10\% training data, the full version `Ours' produces 40.9\% F$_1$ score, significantly outperforming the version without first-stage training (27.6\% F$_1$ score). It is even higher than the version without first-stage training using 50\% labeled training data (39.8\% F$_1$ score). Similar results can be seen on the JRDB-Group benchmark.
				This demonstrates the effectiveness of the proposed self-supervised first-stage training strategy, which can reduce the dependence of the deep network on large-scale annotated training data.
				We can also see that `Ours w/o fine-tune.' produces promising results when using a small amount of training data, e.g., 10\% and 30\%. This is because this setting without training the first-stage representation network significantly reduces the network parameters to be learned.
%							\vspace{-10pt}	
				\begin{table*}[htbp]
					\caption{Performance analysis using different amount of training data on PANDA and JRDB-Group datasets (\%).} %	\vspace{-10pt}
					\label{tab:traindata} 
					\begin{spacing}{1.05}
						\centering
						\renewcommand\tabcolsep{5pt}
						\footnotesize
						\begin{tabular}[*c]{c|c|ccc|ccc}
							\Xhline{1pt}
							%\rowcolor{mygray}  
							& &\multicolumn{3}{c|}{PANDA}  & \multicolumn{3}{c}{JRDB-Group} \\ \cline{3-8} 
							\multirow{-2}{30pt}{Method} &\multirow{-2}{20pt}{Data} &{Precision} & {Recall} & {F1}  &{Precision} & {Recall} & {F1}\\  \cline{2-4} \hline
							% \rowcolor{mygray} \hline
							\multirow{4}{60pt}{\\w/o 1$^{st}$-train} &10\%	Train. 	&37.7	&30.7	&33.8	&40.5 &22.7 &29.1 \\
							\rowcolor{mygray}
							\cellcolor{white} Ours &30\%	Train. 	      &41.9	&34.7	&38.0	& 39.1 & 46.0 & 42.2 \\
							&50\%	Train.  		&49.1	&37.3	&42.2	& 37.5& 48.8& 42.4 \\\cline{2-8} 
							\rowcolor{mygray}
							\cellcolor{white} &100\%  Train.        &48.4	&41.3	&44.6 &49.8 &47.8 &48.8 \\
							\hline   \hline
							\multirow{4}{60pt}{\\ w/o fine-tune.} &10\%	Train.  &61.7	&38.7	&47.5	&38.4 &50.9 &43.8\\
							\rowcolor{mygray}
							\cellcolor{white} Ours &30\%	Train. 	   &54.1	&44.0	&48.5	& 51.6 &44.0 &47.5 \\
							&50\%	Train.  	&54.8	&45.3	&49.6	& 49.1 & 45.5 & 47.3 \\ \cline{2-8} 
							\rowcolor{mygray}
							\cellcolor{white}	&100\%  Train.      &58.6	&45.3	&51.1 &50.3 &48.1 &49.2 \\
							\hline   \hline
							&10\%	Train. 	&43.5	&36.0	&39.4	&42.2  &46.4  &44.2  \\
							\rowcolor{mygray}
							\cellcolor{white} &30\%	Train. 	 &46.9	&40.0	&43.2 & 52.9 &42.8 &47.3  \\
							&50\%	Train.  &55.1	&50.7	&52.8  &53.9 &42.4 &47.4  \\\cline{2-8} 
							\rowcolor{mygray}
							\cellcolor{white} \multirow{-4}{25pt}{Ours} &100\%	Train.   &55.9	&50.7	&53.2 	& 57.7 & 56.2 & 56.9 \\
							\Xhline{1pt}            
						\end{tabular}
					\end{spacing} %\vspace{-10pt}
				\end{table*}

				We further provide the group detection results using two new metrics.
				The commonly-used Half metric~\cite{choi2014discovering} takes a fixed threshold of 0.5 of group IOU for evaluating the group detection accuracy.
				In some cases, the threshold of 0.5 is a little loose, and a single threshold may bring in bias.
				This way, we vary the group IOU threshold of 0.5 in the classical Half metric to larger values until 1 and then calculate the corresponding group detection performance. We plot the curve using the abscissa axis as the threshold, and the vertical axis is the F$_1$ score under each threshold. 
				Then we compute the AUC (Area Under The Curve) score as the metric for evaluation.

				As a harsher metric, we also calculate the matrix IOU of the predicted group relation matrix $\hat{\mathbf{R}}$ and the ground-truth ${\mathbf{R}}_\mathrm{gt}$ by 
				$
				\mathrm{IOU^{GM}} = \frac{\sum \mathrm{AND}(\hat{\textbf{R}},{\textbf{R}_\mathrm{gt}})}{\sum \mathrm{OR}(\hat{\textbf{R}},{\textbf{R}_\mathrm{gt}})}
				$,
				where $\mathrm{AND}$ and $\mathrm{OR}$ denotes the logical operations functions, and $\sum$ denotes the summation over all values. The group relation metric encodes the group relations between each pair of subjects, and the metric $\mathrm{IOU^{GM}}$ evaluates the pair-wise group relation accuracy among all subjects.
				
%				\vspace{-10pt}
				\begin{table}[htbp] 
					\caption{Comparative group detection results using IOU-AUC and $\mathrm{IOU^{GM}}$ (\%).} %	\vspace{-10pt}
					\label{tab:groupIOU} 
					\begin{spacing}{1.05}
						\centering
						\renewcommand\tabcolsep{8pt}
						\footnotesize
						\begin{tabular}[*c]{l|cc|cc}
							\Xhline{1pt}
							%\rowcolor{mygray}  
							&\multicolumn{2}{c|}{PANDA} &	\multicolumn{2}{c}{JRDB-Group} \\ \cline{2-5}
							&{IOU-AUC} &$\mathrm{IOU^{GM}}$  &{IOU-AUC} &$\mathrm{IOU^{GM}}$ \\  \cline{1-5}
							Dis.Mat + \cite{zhan2018consensus} 	&17.0 	&5.9  & 14.1  &12.9 \\	
							\rowcolor{mygray}				
							GNN w GRU		&16.3 	&6.5  & 21.7 & 20.1	  \\
							ARG~\cite{GAR-ARG}	&19.3 		&6.4  	&21.6 & 19.3\\
							%			Social STGCN    	& 0.0		& 0.0 	& 0.0 \\
							\hline 
							\rowcolor{mygray}
							Joint~\cite{ehsanpour2020joint}  &-  &-  & 20.4 & 16.6  \\
							%						\rowcolor{mygray}
							JRDB-Group~\cite{Ehsanpour2021JRDBActAL}  &-  &-  & 26.3 & 20.6 \\
							\hline   \hline
							%			Ours w  only FC         &50.3 &48.1 &49.2 \\
							%			Ours w  one-stage train. &49.8 &47.8 &48.8 \\
							\rowcolor{mygray}
							Ours			&41.2 & 31.2 &42.5  &32.5\\
							\Xhline{1pt}            
						\end{tabular}
					\end{spacing} % \vspace{-0pt}
				\end{table}
				
				We show the results using the new metrics in Table~\ref{tab:groupIOU}. Note that, on the PANDA dataset, we do not conduct the experiments using the methods of Joint and JRDB-Group, since their deep CNN features can not be applied to the scene with thousands of people in the PANDA dataset. We can not provide the evaluation of new metrics for the baseline methods in PANDA, i.e., `Group-to-local', because it did not provide the source code and raw results.
				From Table~\ref{tab:groupIOU} we can first see that our method shows remarkable performance compared to the other methods.
				Specifically, on the JRDB dataset under the metrics of IOU-AUC and $\mathrm{IOU^{GM}}$, the simple baseline method using the distance metric based method can not provide comparable or better results than other methods like that in Table~\ref{tab:jrdb}. 
				%		This reflects the rationality of the proposed metrics to a degree.
%				\ljc{
				We can also see that the performance under $\mathrm{IOU^{GM}}$ is generally low.
%				especially on the PANDA dataset.} This is mainly because of the mass number of people in the scene. 
			    This also leaves space for the development of more effective algorithms.
				
				\section{Conclusion}
				
				In this paper, we have proposed a new framework for social group detection in large-scale multi-person scenes. We developed a double-head two-stage network for the human relation representation and group detection, in which we designed a self-supervised first-stage training strategy for relation representation. We evaluated the proposed method on two state-of-the-art benchmarks, i.e., PANDA and JRDB-Group. The proposed method outperforms all the comparison methods by a very large margin. We also verified the effectiveness of the self-supervised representation training strategy by using very few (labeled) training data.
				
				\textbf{Acknowledgment}. This work was supported in part by the National Natural Science Foundation of China under Grants U1803264, 62072334, and the Tianjin Research Innovation Project for Postgraduate Students under Grant 2021YJSB174.
				
				{
					\bibliographystyle{splncs04}
					\bibliography{IEEEabrv,SGD}

\begin{thebibliography}{10}
\providecommand{\url}[1]{\texttt{#1}}
\providecommand{\urlprefix}{URL }
\providecommand{\doi}[1]{https://doi.org/#1}

\bibitem{MPII}
Andriluka, M., Pishchulin, L., Gehler, P., Schiele, B.: 2d human pose
  estimation: New benchmark and state of the art analysis. In: IEEE Conference
  on Computer Vision and Pattern Recognition (2014)

\bibitem{SSU}
Bagautdinov, T., Alahi, A., Fleuret, F., Fua, P., Savarese, S.: Social scene
  understanding: End-to-end multi-person action localization and collective
  activity recognition. In: IEEE Conference on Computer Vision and Pattern
  Recognition (2017)

\bibitem{bazzani2012decentralized}
Bazzani, L., Cristani, M., Murino, V.: Decentralized particle filter for joint
  individual-group tracking. In: IEEE Conference on Computer Vision and Pattern
  Recognition (2012)

\bibitem{chang2011probabilistic}
Chang, M.C., Krahnstoever, N., Ge, W.: Probabilistic group-level motion
  analysis and scenario recognition. In: IEEE International Conference on
  Computer Vision (2011)

\bibitem{shiftgcn}
Cheng, K., Zhang, Y., He, X., Chen, W., Cheng, J., Lu, H.: Skeleton-based
  action recognition with shift graph convolutional network. In: IEEE/CVF
  Conference on Computer Vision and Pattern Recognition (2020)

\bibitem{choi2014discovering}
Choi, W., Chao, Y.W., Pantofaru, C., Savarese, S.: Discovering groups of people
  in images. In: European Conference on Computer Vision (2014)

\bibitem{ehsanpour2020joint}
Ehsanpour, M., Abedin, A., Saleh, F., Shi, J., Reid, I., Rezatofighi, H.: Joint
  learning of social groups, individuals action and sub-group activities in
  videos. In: European Conference on Computer Vision (2020)

\bibitem{ehsanpour2021jrdb}
Ehsanpour, M., Saleh, F., Savarese, S., Reid, I., Rezatofighi, H.: Jrdb-act: A
  large-scale multi-modal dataset for spatio-temporal action, social group and
  activity detection. arXiv Preprint arXiv:2106.08827  (2021)

\bibitem{Ehsanpour2021JRDBActAL}
Ehsanpour, M., Saleh, F.S., Savarese, S., Reid, I.D., Rezatofighi, H.:
  Jrdb-act: A large-scale dataset for spatio-temporal action, social group and
  activity detection (2021)

\bibitem{Fan_2019_ICCV}
Fan, L., Wang, W., Huang, S., Tang, X., Zhu, S.C.: Understanding human gaze
  communication by spatio-temporal graph reasoning. In: IEEE/CVF International
  Conference on Computer Vision (2019)

\bibitem{feldmann2010tracking}
Feldmann, M., Fr{\"a}nken, D., Koch, W.: Tracking of extended objects and group
  targets using random matrices. IEEE Transactions on Signal Processing
  \textbf{59}(4),  1409--1420 (2010)

\bibitem{fernando2018gd}
Fernando, T., Denman, S., Sridharan, S., Fookes, C.: Gd-gan: Generative
  adversarial networks for trajectory prediction and group detection in crowds.
  In: Asian Conference on Computer Vision (2018)

\bibitem{gan2021self}
Gan, Y., Han, R., Yin, L., Feng, W., Wang, S.: Self-supervised multi-view
  multi-human association and tracking. In: ACM International Conference on
  Multimedia (2021)

\bibitem{AT}
Gavrilyuk, K., Sanford, R., Javan, M., Snoek, C.G.: Actor-transformers for
  group activity recognition. In: IEEE/CVF Conference on Computer Vision and
  Pattern Recognition (2020)

\bibitem{ge2012vision}
Ge, W., Collins, R.T., Ruback, R.B.: Vision-based analysis of small groups in
  pedestrian crowds. IEEE Transactions on Pattern Analysis and Machine
  Intelligence  \textbf{34}(5),  1003--1016 (2012)

\bibitem{Goel_2019_CVPR}
Goel, A., Ma, K.T., Tan, C.: An end-to-end network for generating social
  relationship graphs. In: Proceedings of the IEEE/CVF Conference on Computer
  Vision and Pattern Recognition (2019)

\bibitem{gupta2018social}
Gupta, A., Johnson, J., Fei-Fei, L., Savarese, S., Alahi, A.: Social gan:
  Socially acceptable trajectories with generative adversarial networks. In:
  IEEE Conference on Computer Vision and Pattern Recognition (2018)

\bibitem{han2021multiple}
Han, R., Feng, W., Zhang, Y., Zhao, J., Wang, S.: Multiple human association
  and tracking from egocentric and complementary top views. IEEE TPAMI  (2021).
  \doi{10.1109/TPAMI.2021.3070562}

\bibitem{han2020complementary}
Han, R., Feng, W., Zhao, J., Niu, Z., Zhang, Y., Wan, L., Wang, S.:
  Complementary-view multiple human tracking. In: AAAI Conference on Artificial
  Intelligence (2020)

\bibitem{Han_2022_CVPR}
Han, R., Gan, Y., Li, J., Wang, F., Feng, W., Wang, S.: Connecting the
  complementary-view videos: Joint camera identification and subject
  association. In: IEEE/CVF Conference on Computer Vision and Pattern
  Recognition (2022)

\bibitem{han2022multiview}
Han, R., Wang, Y., Yan, H., Feng, W., Wang, S.: Multi-view multi-human
  association with deep assignment network. IEEE TIP  \textbf{31},  1830--1840
  (2022)

\bibitem{lerner2007crowds}
Lerner, A., Chrysanthou, Y., Lischinski, D.: Crowds by example. In: Computer
  Graphics Forum (2007)

\bibitem{li2020graph}
Li, W., Duan, Y., Lu, J., Feng, J., Zhou, J.: Graph-based social relation
  reasoning. In: European Conference on Computer Vision (2020)

\bibitem{martin2021jrdb}
Mart{\'\i}n-Mart{\'\i}n, R., Patel, M., Rezatofighi, H., Shenoi, A., Gwak, J.,
  Frankel, E., Sadeghian, A., Savarese, S.: Jrdb: A dataset and benchmark of
  egocentric robot visual perception of humans in built environments. IEEE
  Transactions on Pattern Analysis and Machine Intelligence  (2021)

\bibitem{mohamed2020social}
Mohamed, A., Qian, K., Elhoseiny, M., Claudel, C.: Social-stgcnn: A social
  spatio-temporal graph convolutional neural network for human trajectory
  prediction. In: IEEE/CVF Conference on Computer Vision and Pattern
  Recognition (2020)

\bibitem{monfort2019moments}
Monfort, M., Andonian, A., Zhou, B., Ramakrishnan, K., Bargal, S.A., Yan, T.,
  Brown, L., Fan, Q., Gutfreund, D., Vondrick, C., et~al.: Moments in time
  dataset: one million videos for event understanding. IEEE Transactions on
  Pattern Analysis and Machine Intelligence  \textbf{42}(2),  502--508 (2019)

\bibitem{moussaid2010walking}
Moussa{\"\i}d, M., Perozo, N., Garnier, S., Helbing, D., Theraulaz, G.: The
  walking behaviour of pedestrian social groups and its impact on crowd
  dynamics. PloS one  \textbf{5}(4),  e10047 (2010)

\bibitem{pang2011detection}
Pang, S.K., Li, J., Godsill, S.J.: Detection and tracking of coordinated
  groups. IEEE Transactions on Aerospace and Electronic Systems
  \textbf{47}(1),  472--502 (2011)

\bibitem{pellegrini2009you}
Pellegrini, S., Ess, A., Schindler, K., Van~Gool, L.: You'll never walk alone:
  Modeling social behavior for multi-target tracking. In: IEEE International
  Conference on Computer Vision (2009)

\bibitem{pellegrini2010improving}
Pellegrini, S., Ess, A., Van~Gool, L.: Improving data association by joint
  modeling of pedestrian trajectories and groupings. In: European Conference on
  Computer Vision (2010)

\bibitem{pramono2020empowering}
Pramono, R.R.A., Chen, Y.T., Fang, W.H.: Empowering relational network by
  self-attention augmented conditional random fields for group activity
  recognition. In: European Conference on Computer Vision (2020)

\bibitem{shao2018real}
Shao, J., Dong, N., Zhao, Q.: A real-time algorithm for small group detection
  in medium density crowds. Pattern Recognition and Image Analysis
  \textbf{28}(2),  282--287 (2018)

\bibitem{shao2014scene}
Shao, J., Change~Loy, C., Wang, X.: Scene-independent group profiling in crowd.
  In: IEEE Conference on Computer Vision and Pattern Recognition (2014)

\bibitem{solera2015socially}
Solera, F., Calderara, S., Cucchiara, R.: Socially constrained structural
  learning for groups detection in crowd. IEEE Transactions on Pattern Analysis
  and Machine Intelligence  \textbf{38}(5),  995--1008 (2015)

\bibitem{swofford2020improving}
Swofford, M., Peruzzi, J., Tsoi, N., Thompson, S., Mart{\'\i}n-Mart{\'\i}n, R.,
  Savarese, S., V{\'a}zquez, M.: Improving social awareness through dante: Deep
  affinity network for clustering conversational interactants. ACM on
  Human-Computer Interaction  \textbf{4}(CSCW1),  1--23 (2020)

\bibitem{thompson2021conversational}
Thompson, S., Gupta, A., Gupta, A.W., Chen, A., V{\'a}zquez, M.: Conversational
  group detection with graph neural networks. In: International Conference on
  Multimodal Interaction (2021)

\bibitem{turner2010towards}
Turner, J.C.: Towards a cognitive redefinition of the social group. In:
  Research Colloquium on Social Identity of the European Laboratory of Social
  Psychology. Psychology Press (2010)

\bibitem{vaswani2017attention}
Vaswani, A., Shazeer, N., Parmar, N., Uszkoreit, J., Jones, L., Gomez, A.N.,
  Kaiser, {\L}., Polosukhin, I.: Attention is all you need. In: Advances in
  Neural Information Processing Systems (2017)

\bibitem{wang2020panda}
Wang, X., Zhang, X., Zhu, Y., Guo, Y., Yuan, X., Xiang, L., Wang, Z., Ding, G.,
  Brady, D., Dai, Q., et~al.: Panda: A gigapixel-level human-centric video
  dataset. In: IEEE/CVF Conference on Computer Vision and Pattern Recognition
  (2020)

\bibitem{GAR-ARG}
Wu, J., Wang, L., Wang, L., Guo, J., Wu, G.: Learning actor relation graphs for
  group activity recognition. In: IEEE/CVF Conference on Computer Vision and
  Pattern Recognition (2019)

\bibitem{yamaguchi2011you}
Yamaguchi, K., Berg, A.C., Ortiz, L.E., Berg, T.L.: Who are you with and where
  are you going? In: IEEE Conference on Computer Vision and Pattern Recognition
  (2011)

\bibitem{dataset-NBA}
Yan, R., Xie, L., Tang, J., Shu, X., Tian, Q.: Social adaptive module for
  weakly-supervised group activity recognition. In: European Conference on
  Computer Vision (2020)

\bibitem{yuan2021learning}
Yuan, H., Ni, D.: Learning visual context for group activity recognition. In:
  AAAI Conference on Artificial Intelligence (2021)

\bibitem{zhan2018consensus}
Zhan, X., Liu, Z., Yan, J., Lin, D., Loy, C.C.: Consensus-driven propagation in
  massive unlabeled data for face recognition. In: European Conference on
  Computer Vision (2018)

\bibitem{zhao2020human}
Zhao, J., Han, R., Gan, Y., Wan, L., Feng, W., Wang, S.: Human identification
  and interaction detection in cross-view multi-person videos with wearable
  cameras. In: ACM International Conference on Multimedia (2020)

\bibitem{Zhou_2020_CVPR}
Zhou, T., Wang, W., Qi, S., Ling, H., Shen, J.: Cascaded human-object
  interaction recognition. In: IEEE/CVF Conference on Computer Vision and
  Pattern Recognition (2020)

\bibitem{Zou_2021_CVPR}
Zou, C., Wang, B., Hu, Y., Liu, J., Wu, Q., Zhao, Y., Li, B., Zhang, C., Zhang,
  C., Wei, Y., Sun, J.: End-to-end human object interaction detection with hoi
  transformer. In: IEEE/CVF Conference on Computer Vision and Pattern
  Recognition (2021)

\end{thebibliography}
				}
				
			\end{document}